\def\year{2020}\relax
\documentclass[letterpaper]{article} % DO NOT CHANGE THIS
\usepackage{aaai20}  % DO NOT CHANGE THIS
\usepackage{times}  % DO NOT CHANGE THIS
\usepackage{helvet} % DO NOT CHANGE THIS
\usepackage{courier}  % DO NOT CHANGE THIS
\usepackage[hyphens]{url}  % DO NOT CHANGE THIS
\usepackage{graphicx} % DO NOT CHANGE THIS
\usepackage{graphicx}  % DO NOT CHANGE THIS
\usepackage{multirow}
\usepackage{algorithm}
\usepackage{algorithmic}
\usepackage{amsfonts}
\usepackage{amsmath}
\usepackage{subfigure}
\usepackage{color}
\usepackage[hidelinks]{hyperref}
\title{Hadamard Codebook Based Deep Hashing}
\author{
	Shen Chen\textsuperscript{\rm 1},
	Liujuan Cao\textsuperscript{\rm 1},
	Mingbao Lin\textsuperscript{\rm 1},
	Yan Wang\textsuperscript{\rm 2},
	Xiaoshuai Sun\textsuperscript{\rm 1},\\
	\Large \textbf{Chenglin Wu\textsuperscript{\rm 3},
		Jingfei Qiu\textsuperscript{\rm 4}
		\and Rongrong Ji\textsuperscript{\rm 1,4}\thanks{Corresponding Author.}}\\
	\textsuperscript{\rm 1}Fujian Laboratory of Sensing and Computing for Smart City,\\
	Department Cognitive Science, School of Informatics, Xiamen University, China,\\
	\textsuperscript{\rm 2}Pinterest, San Francisco, USA,
	\textsuperscript{\rm 3}Fuzhi, Xiamen, China,
	\textsuperscript{\rm 4}Peng Cheng Laboratory, Shenzhen, China\\
	\{chenshen,lmbxmu\}@stu.xmu.edu.cn, \{caoliujuan,xssun,rrji\}@xmu.edu.cn,\\
	yanw@pinterest.com, alexanderwu@fuzhi.ai, qiujf@pcl.ac.cn}
\begin{document}
	
	\maketitle
	
	\begin{abstract}
		As an approximate nearest neighbor search technique, hashing has been widely applied in large-scale image retrieval due to its excellent efficiency.
        Most supervised deep hashing methods have similar loss designs with embedding learning, while quantizing the continuous high-dim feature into compact binary space.
        We argue that the existing deep hashing schemes are defective in two issues that seriously affect the performance, \emph{i.e.}, bit independence and bit balance.
        The former refers to hash codes of different classes should be independent of each other, while the latter means each bit should have a balanced distribution of $+1s$ and $-1s$.
        In this paper, we propose a novel supervised deep hashing method, termed Hadamard Codebook based Deep Hashing (HCDH), which solves the above two problems in a unified formulation.
        Specifically, we utilize an off-the-shelf algorithm to generate a binary Hadamard codebook to satisfy the requirement of bit independence and bit balance, which subsequently serves as the desired outputs of the hash functions learning.
       	We also introduce a projection matrix to solve the inconsistency between the order of Hadamard matrix and the number of classes.
       	Besides, the proposed HCDH further exploits the supervised labels by constructing a classifier on top of the outputs of hash functions.
        Extensive experiments demonstrate that HCDH can yield discriminative and balanced binary codes, which well outperforms many state-of-the-arts on three widely-used benchmarks.
	\end{abstract}
	
	\section{Introduction}
	With the rapid growth of image data on the Internet, approximate nearest neighbor (ANN) search has attracted extensive research attention.
	Among various ANN techniques, hashing has been a popular solution due to the low storage cost and fast retrieval speed~\cite{Gionis1999SimilaritySI,Weiss2008SpectralH,Liu2011HashingWG,Liu2012SupervisedHW,Gong2013IterativeQA,Xia2014SupervisedHF,Yang2015SupervisedLO,Li2015FeatureLB,Li2017DeepSD,cao2017hashnet,Yang2018,cakir2019hashing}.
    Hashing aims to transform high-dimensional continuous feature into compact binary codes, while preserving the structure of the original data.
    Coming with the recent advance in deep learning, the recent trend of hashing has focused on leveraging deep models to generate hash codes~\cite{Xia2014SupervisedHF,Lai2015SimultaneousFL,Li2015FeatureLB,cao2017hashnet,cakir2019hashing}, which have shown superior improvements over the traditional hashing methods like Locality Sensitive Hashing (LSH)~\cite{Gionis1999SimilaritySI}, Spectral Hashing (SH)~\cite{Weiss2008SpectralH} and Iterative Quantization (ITQ)~\cite{Gong2013IterativeQA}.
    
    One practical challenge in hashing is the binary constraint. To solve the challenge, most state-of-the-art deep hashing methods~\cite{zhu2016deep,cao2017hashnet,cakir2019hashing} follow a very similar design as \textit{embedding learning}, with tweaks on the binary constraint and optimization techniques.
    For instance, HashNet~\cite{cao2017hashnet} preserves similarity information of pairwise images by weighted maximum likelihood.
    MIHash~\cite{cakir2019hashing} optimizes the Mutual Information~\cite{cover2012elements} of neighbors and non-neighbors.
    However, there exist an inherent resemblance between hashing and embedding learning. And it retains unclear whether such resemblance is due to that the discrete variable constraints do not change the problem structure, or indeed overlook, the intrinsic problem of hash function learning. 
    
    %%%%%%%%%%%%%%%%%%%%%%%%%%%%%%%%%%%
    \begin{figure*}[t]
    	\begin{center}
    		\includegraphics[width=1\textwidth]{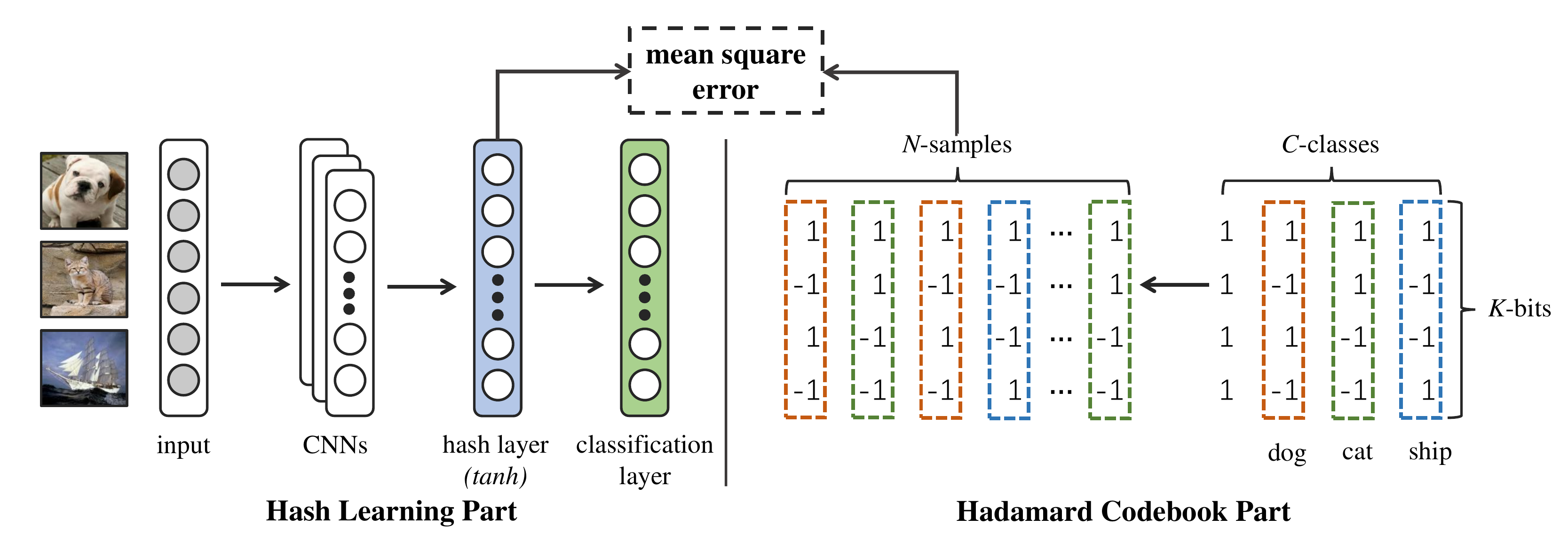}
    	\end{center}
    	\caption{\label{network}
       		The framework of the proposed Hadamard Codebook Based Deep Hash (HCDH). In the Hadamard Codebook Part, we first generate an independent and balanced Hadamard matrix via an off-the-shelf algorithm. Then $C$ column vectors from Hadamard matrix are randomly selected to form Hadamard codebook, which serves as the anchor to guide the learning of hash codes. In the Hash Learning Part, we construct a hash layer with $K$ units and a classification layer with $C$ units on top of the feature layer of the pre-trained CNNs. The hash layer adopt \textit{tanh} activation function to approximate \textit{sign} function. HCDH takes input from point-wise data and learns image representation, binary codes, and classification through the optimization of mean square error and classification error. 
    	}
    \end{figure*}
    %%%%%%%%%%%%%%%%%%%%%%%%%%%%%%%%%%%	
    
    To take a deeper look at the intrinsic problem, following the loss design tricks scattered in the traditional hashing methods~\cite{Weiss2008SpectralH,liu2010large,Liu2011HashingWG,liu2014discrete}, we argue that two important properties are long undermined in the existing deep hashing methods, \emph{i.e.}, \textit{bit independence} and \textit{bit balance}.
    In terms of bit independence, hash codes of different classes should be independent of each other, which can be interpreted from the information theory perspective.
    Under a setting of fixed dimensions, independent hash codes (whether it is from random projection or by design) could take better advantage of the hash bits, which is also validated in ~\cite{Gong2013IterativeQA}.
    In terms of bit imbalance, the values of a bit should not be sparsely distributed, \emph{i.e.}, mostly $1$ or $-1$, but should instead have a balanced distribution.
    As a direct validation, instead of directly using a sign function with a threshold to quantize a hash bit (either 1 or -1), we simply replace $sgn(x)$ with $sgn(x-\text{mean})$ and achieve 1.9\% $m$AP gain for HashNet model on CIFAR-10~\cite{krizhevsky2009learning}.
    Note that some works~\cite{Yang2015SupervisedLO,shen2017deep} proposed to achieve these two properties by introducing \textit{independence loss} and \textit{balance loss}, the performance of which is however limited as the numerical optimization often leads to a local minimum and also inevitably introduces more hyper-parameters to be tuned.
    Although such loss considers reasonable expectation on the hash codes, the optimization may be still suboptimal by lacking analytical guidance from quantized statistics such as codebooks to guide the search.
    
	In this paper, inspired by previous works~\cite{Lin2018SupervisedOH,Lin2019HadamardMG} that primarily addressed online hashing, we propose a Hadamard Codebook based Deep Hashing (HCDH), which tackles the above challenges in a unified framework. 
	In principle, we resort to the power tool of Hadamard matrix that is absolutely independent and balanced (with a tiny cost of one wasted bit), together with the recent advances in deep feature learning, as illustrated in Fig.\,\ref{network}.
    In the training stage, we generate the Hadamard matrix with $K$ bits via an off-the-shelf algorithm~\cite{sylvester1867lx}.
    Then $C$ column vectors are randomly and non-repeatedly selected from Hadamard matrix to serve as the anchor codebook, referred as \textit{Hadamard codebook}, which guides the learning of hash codes.
    In a supervised setting, we also introduce a projection matrix to solve the inconsistency between the order of Hadamard matrix and the number of classes.
    Under such a circumstance, the learned hash codes have a good separation between classes, with maximized information gain for each bit.
    To further exploit the supervised labels, we incorporate a deep classifier into the binary code learning process, as illustrated in the Hash Learning part of Fig.\,\ref{network}, which enables the discriminative hash codes and image representation can be learned simultaneously in a scalable end-to-end fashion.
	
	Our main contributions can be summarized as follows:
	
	\begin{itemize}
		\item We identify two key issues that are long overlooked in the existing deep hashing methods, \emph{i.e.}, bit independence and bit balance.
		\item We propose a unified Hadamard codebook model together with deep feature learning to tackle the above two issues. The proposed HCDH method is robust, efficient and scalable for large-scale image retrieval.
		\item Extensive experiments demonstrate that the proposed HCDH well beats the existing state-of-the-arts~\cite{Gionis1999SimilaritySI,Gong2013IterativeQA,Liu2012SupervisedHW,Shen2015SupervisedDH,Yang2015SupervisedLO,Li2015FeatureLB,Li2017DeepSD,cao2017hashnet,cakir2019hashing} on three widely-used benchmark datasets, \emph{i.e.}, CIFAR-10, NUS-WIDE, and ImageNet. 
	\end{itemize}
	
	\section{The Proposed Approach}
	
	\subsection{Problem Definition}
	Let $\mathbf{X}=\left\{\mathbf{x}_i\right\}_{i=1}^{N}$ denote a set of $N$ training images labeled with $C$ classes. Each image belongs to one class (single-label case) or several classes (multi-label case). Without loss of generality, we consider a label matrix $\mathbf{Y}=\left\{\mathbf{y}_i\right\}_{i=1}^{N}$, where $\mathbf{y}_i\in\{0,1\}^{C}$ denotes the label encoding of $\mathbf{x}_i$, and $C$ is the number of classes. The $c$-th element of $\mathbf{y}_i$ being 1 indicates that $\mathbf{x}_i$ belongs to class $c$.
	The goal of hashing is to learn a mapping $\Omega:\mathbf{X} \rightarrow\{-1,1\}^{K \times N}$ that projects input points $\mathbf{X}$ into $K$-bit compact hash codes $\mathbf{B}=\Omega\left(\mathbf{X}\right)$.
	
	\subsection{The Framework}
	Unlike previous works that explicit model bit independence and balance in the loss design~\cite{Yang2015SupervisedLO,shen2017deep}, we aim to find a projection matrix $\mathbf{W}\in\mathbb{R}^{C \times K}$ that transforms the label matrix $\mathbf{Y}$ from the label space to the Hamming space, in which the bit balance and bit independence are well preserved.
	$\mathbf{Y}$ in Hamming space then serves as anchors to guide the learning of hash codes $\mathbf{B}$. That is:
	\begin{equation}
	\label{loss1}
	\begin{aligned} \min_{\mathbf{W},\Omega} \ &\mathit{L} = \frac{1}{2}||\mathbf{W}^T \mathbf{Y}-\mathbf{B}\|^{2} \\ &\mathit{s.t.} \ \mathbf{B}=\Omega\left(\mathbf{X}\right). \end{aligned}
	\end{equation}
	
	The above optimization depends on both matrix $\mathbf{W}$ and mapping $\Omega$, which correspond to the Hadamard Codebook and the Hash Learning, as shown in Fig.\,\ref{network}. In the following, we show that the optimal matrix \textbf{W} can be obtained directly rather than learning, and the mapping $\Omega$ is learned via deep neural network.
	
	\subsection{Hadamard Codebook}
	As stated in~\cite{Weiss2008SpectralH}, the optimal hash codes $\mathbf{B}$ should satisfy: 1) Independence: Hash codes of different classes are independent of each other. 2) Balance: Each bit has a 50\% chance of being 1 or -1. We formulate the definition of bit independence and bit balance by the following:
	\begin{equation}
	\label{independence}
	\mathbf{W}^{T} \mathbf{W}=\mathbf{I},
	\end{equation}
	\begin{equation}
	\label{balance}
	\mathbf{W}^{T} \mathbf{1}=\mathbf{0}.
	\end{equation}

	We adopt Hadamard matrix~\cite{hadamard1893resolution} as the backbone to construct the codebook of classes, which well conforms the properties of independence and balance.
	Specifically, the Hadamard matrix is an $n$-order orthogonal matrix, \emph{i.e.}, both its row vectors and column vectors are pairwisely orthogonal, which by nature satisfies Eq.\,(\ref{independence}). In other words:
	\begin{equation}
	\mathbf{H H}^{T}=n \mathbf{I}_{n}, \text { or } \mathbf{H}^{T} \mathbf{H}=n \mathbf{I}_{n},
	\end{equation}
	where $\mathbf{H}$ is a Hadamard matrix and $\mathbf{I}_{n}$ is an $n$-order identity matrix. Besides, rows or columns of $\mathbf{H}$ are half $+1s$ and half $-1s$ (except the first row or column), which by nature satisfies Eq.\,(\ref{balance}).
	
	Hence, by eliminating the first row or the first column, Hadamard matrix can be used as an efficient codebook for learning hash codes (referred \textit{Hadamard codebook}).
	
	Practically, a $2^k$-order Hadamard matrix can be constructed recursively by the Sylvester’s algorithm~\cite{sylvester1867lx}. That is:
	\begin{equation}
	\label{generate-hadamard}
	\begin{aligned} \mathbf{H}_{2^{k}} & =\left[\begin{array}{cc}{\mathbf{H}_{2^{k-1}}} & {\mathbf{H}_{2^{k-1}}} \\ {\mathbf{H}_{2^{k-1}}} & {-\mathbf{H}_{2^{k-1}}}\end{array}\right], \\ \mathbf{H}_{2} & =\left[\begin{array}{rr}{1} & {1} \\ {1} & {-1}\end{array}\right].\end{aligned} 
	\end{equation}
	
	Furthermore, Hadamard matrix of orders 12 and 20 can be constructed by Hadamard transformation~\cite{hadamard1893resolution}. Since code length frequently used in applications of binary hashing is $2^k$, we mainly adopt Eq.\,(\ref{generate-hadamard}) to generate the target binary codes.
	
	Therefore, given the bit number $K$ and the classes number $C$, we generate the $K$-order Hadamard matrix by Eq.\,(\ref{generate-hadamard}), and then randomly select $C$ column vectors as the Hadamard codebook.
	Each column vector of Hadamard codebook serves as a unique codeword for each class.
	However, the key problem is that this is only feasible when $C$ is less than the bit number $K$, which disobeys the real-world scenario where the class number $C$ is often much larger than $K$. Namely, for the case of $C > K$, there are not enough column vectors in Hadamard codebook to assign a unique codeword for each class, making it impossible to ensure the generated Hadamard codebook are orthogonal to each other. To solve the above problem, we define the order of Hadamard matrix $K^{*}$ as follows:
	\begin{equation}
	\label{hadamard-order}
	K^{*}=\min \{r|r=2^{k}, r \geq K, r \geq C, k=1,2, \dots\}.
	\end{equation}
	
	To further solve the inconsistency between $K^*$ and $K$, we utilize LSH~\cite{Gionis1999SimilaritySI} to randomly generate a Gaussian distribution matrix $\mathbf{T} \in \mathbb{R}^{K^* \times K}$, which transforms Hadamard matrix $\mathbf{H^*}$ from $\mathbb{R}^{K^* \times K^*}$ to $\mathbb{R}^{K^* \times K}$. Furthermore, the sign function is adopted to obtain the desired binary codes as:
	\begin{equation}
	\label{hadamard-case2}
	\mathbf{H^*}=sgn(\mathbf{H^*T}).
	\end{equation}
	
	Finally, $C$ column vectors from matrix $\mathbf{H^*}$ are randomly and non-repeatedly selected to form the Hadamard codebook $\mathbf{H}$, which serves as the anchor codebook to guide the learning of hash codes, as elaborated later.
	
	So far, we have got the Hadamard codebook that satisfies the two properties defined by Eq.\,(\ref{independence}) and Eq.\,(\ref{balance}). We then reformulate Eq.\,(\ref{loss1}) and define the \textit{hadamad loss} as:
	\begin{equation}
	\label{hadamard-loss}
	\begin{aligned}	\min_{\Omega} \ &\mathit{L}_{H}=\frac{1}{2}||\mathbf{H}^T\mathbf{Y}-\mathbf{B}\|^{2} \\ &\mathit{s.t.} \ \mathbf{B}=\Omega\left(\mathbf{X}\right). \end{aligned}
	\end{equation}
	
	The generation for the desired Hadamard codebook is summarized in Alg. \ref{alg1}.
	
	%%%%%%%%%%%%%%%%%%%%%%%%%%%%%%%%%%%
	\begin{algorithm}[t]
		\caption{The generation for Hadamard Codebook \label{alg1}}
		\renewcommand{\algorithmicrequire}{\textbf{Input:}}
		\renewcommand{\algorithmicensure}{\textbf{Output:}}
		\begin{algorithmic}[1]
			\REQUIRE
			The number of classes $C$ and code length $K$.
			\ENSURE
			Hadamard Codebook $\mathbf{H} \in \mathbb{R}^{C \times K}$\\
			\STATE Set the value of $K^*$ by Eq.\,(\ref{hadamard-order}).
			\STATE Generate $K^*$-order Hadamard martix $\mathbf{H^*}$ by Eq.\,(\ref{generate-hadamard}).
			\IF {$K^*\not=K$} 
			\STATE Randomly generate $\mathbf{T} \in \mathbb{R}^{K^* \times K}$ from Gaussian distribution.
			\STATE Compute $\mathbf{H^*}$ by Eq.\,(\ref{hadamard-case2}).
			\ENDIF
			\STATE Randomly select $C$ column vectors from matrix $\mathbf{H^*}$ as Hadamard Codebook $\mathbf{H}$.
			\RETURN Hadamard Codebook $\mathbf{H}$.
		\end{algorithmic}
	\end{algorithm}
	%%%%%%%%%%%%%%%%%%%%%%%%%%%%%%%%%%%
	
	\subsection{Learning Hash Functions}
	We further construct the mapping $\Omega$ by adding a hash layer with $K$ units on top of the feature layer of network $\mathcal{F}$, as illustrated in Fig.\,\ref{network}. Accordingly, the hash codes are obtained by taking the sign of the hash layer outputs as:
	\begin{equation}
	\label{deep-hadamard-loss}
	\mathbf{B}=sgn \Big(\mathcal{F}\big(\mathbf{X}; \Theta\big)\Big),
	\end{equation}
	where $\Theta$ denotes the parameters of network $\mathcal{F}$, and $sgn(\cdot)$ is the sign function. Since the sign function $sgn(\cdot)$ makes the problem NP-hard, a soft sign function $tanh(\cdot)$ is adopted as the activation function of hash layer to approximate $sgn(\cdot)$, which brings the new formulation of Eq.\,(\ref{hadamard-loss}) as:
	\begin{equation}
	\label{hadamard-loss-tanh}
	\begin{aligned}	\min_{\Theta} \ &\mathit{L}_{H}=\frac{1}{2}||\mathbf{H}^T\mathbf{Y}-\mathbf{B}\|^{2} \\ &\mathit{s.t.} \ \mathbf{B}=tanh\Big(\mathcal{F}\big(\mathbf{X}; \Theta\big)\Big). \end{aligned}
	\end{equation}
	
	The combination of Hadamard codebook and CNN further enables the linking of hash code to the backend classification tasks.
	In particular, a classification layer is constructed on top of the hash layer.
	Unlike the previous work~\cite{Yang2015SupervisedLO} that treats the classification and code learning into separated streams, we merge both tasks into one-stream to learn simultaneously, \emph{i.e.}, the outputs of hash layer are directly guided by the Hadamard codebook and the deep classifier.
	
	To further improve the adaptability of our approach, we adopt different classification losses for different kinds of labels, \emph{i.e.}, single-label case and multi-label case.
	For the single-label case, we adopt \textit{Cross Entropy Loss} as the classification loss, defined as:
	\begin{equation} 	
	\label{cross-entropy-loss}
	\mathit{L}_{CE}=-\frac{1}{N}\sum_{i=1}^{N} \log \frac{e^{\widetilde{\Theta}_{\mathbf{y}_i}(\mathbf{b}_i)}}{\sum_{j=1}^{n} e^{\widetilde{\Theta}_{\mathbf{y}_j}(\mathbf{b}_i)}},
	\end{equation}
	where $\widetilde{\Theta}$ denotes the parameters of classification layer and $\mathbf{b}_i \in \mathbb{R}^{K}$ denotes the hash layer output of the $i$-th image from the $\mathbf{y}_i$-th class.
	
	For the multi-label case, we adopt \textit{Binary Cross Entropy Loss} as the classification loss, defined as:
	\begin{equation}
	\label{BCE-loss}
	\begin{aligned} \mathit{L}_{BCE}=&-\frac{1}{N C} \sum_{i=1}^{N} \sum_{j=1}^{C}\left({\mathbf{y}_{i j} \cdot \log \frac{e^{\widetilde{\Theta}_{\mathbf{y}_i}(\mathbf{b}_i)}}{\sum_{j=1}^{n} e^{\widetilde{\Theta}_{\mathbf{y}_j}}(\mathbf{b}_i)}}\right.\\ &+\left(1-\mathbf{y}_{i j}\right) \cdot \log \left(1-\frac{e^{\widetilde{\Theta}_{\mathbf{y}_i}(\mathbf{b}_i)}}{\sum_{j=1}^{n} e^{\widetilde{\Theta}_{\mathbf{y}_j}(\mathbf{b}_i)}}\right). \end{aligned}
	\end{equation}
	
	By integrating hadamard loss and classification loss into a unified deep network, we formulate the final optimization problem of HCDH as:
	\begin{equation}
	\label{loss_total}
	\min_{\Theta,\widetilde{\Theta}} \ \mathit{L}_{H} + \lambda \mathit{L}_{CE}, \text{ or } \mathit{L}_{H} + \lambda \mathit{L}_{BCE},	\end{equation}
	where $\lambda$ is a hyper-parameter to balance the hadamard loss and classification loss. The parameters of network, \emph{i.e.}, $\Theta$ and $\widetilde{\Theta}$, are updated via back propagation.
	
	%%%%%%%%%%%%%%%%%%%%%%%%%%%%%%%%%%%
	\begin{table*}[t]
		\centering
		\caption{$m$AP results with respect to different bits number on three datasets.}
		\label{mAP}
		\setlength{\tabcolsep}{5pt}
		\begin{tabular}{c|cccc|cccc|cccc}
			\hline
			\multirow{2}{*}{Method} & \multicolumn{4}{c|}{CIFAR-10} & \multicolumn{4}{c|}{NUS-WIDE} & \multicolumn{4}{c}{ImageNet}\\
			\cline{2-13}
			& 16 bits & 32 bits & 64 bits & 128 bits & 16 bits & 32 bits &64 bits & 128 bits & 16 bits & 32 bits &64 bits & 128 bits \\
			\hline
			LSH				&0.130   &0.146   &0.166   &0.176   &0.475   &0.535   &0.559   &0.629   &0.053   &0.114   &0.174   &0.277\\
			ITQ				&0.179   &0.192   &0.201   &0.215   &0.579   &0.648   &0.682   &0.689   &0.077   &0.180   &0.271   &0.348\\
			KSH 			&0.465   &0.496   &0.517   &0.526   &0.631   &0.639   &0.656   &0.654   &0.241   &0.345   &0.429   &0.472\\
			SDH				&0.483   &0.532   &0.560   &0.565   &0.562   &0.705   &0.713   &0.745   &0.441   &0.550   &0.605   &0.630\\
			SSDH			&0.573   &0.612   &0.685   &0.699   &0.710   &0.763   &0.769   &0.770   &0.527   &0.619   &0.652   &0.686\\ 
			DPSH	 &0.641   &0.659   &0.674   &0.677   &0.767   &0.784   &0.795   &0.808   &0.183   &0.287   &0.384   &0.461\\
			DSDH	 &0.605   &0.623   &0.636   &0.651   &0.778   &0.803   &0.819   &0.828   &0.156   &0.216   &0.282   &0.341\\
			HashNet			&0.663   &0.687   &0.696   &0.705   &\textbf{0.783}   &0.811   &0.829   &\textbf{0.840}   &0.464   &0.593   &0.655   &0.702\\
			MIHash			&\underline{0.690}   &\underline{0.745}   &\underline{0.761}   &\underline{0.773}   &0.760  &0.792   &0.817   &0.829   &\underline{0.565}   &\underline{0.648}   &\underline{0.689}   &\underline{0.708}\\
			HCDH			&\textbf{0.769}   &\textbf{0.774}   &\textbf{0.785}   &\textbf{0.790}  &\underline{0.779}  &\textbf{0.820}   &\textbf{0.830}   &\underline{0.832} &\textbf{0.636}   &\textbf{0.691}   &\textbf{0.719}   &\textbf{0.732}\\
			\hline
		\end{tabular}
		\vspace{10pt}
	\end{table*}
	%%%%%%%%%%%%%%%%%%%%%%%%%%%%%%%%%%%
	
	%%%%%%%%%%%%%%%%%%%%%%%%%%%%%%%%%%%
	\begin{figure*}[t]
		\centering
		\includegraphics[width=1\textwidth]{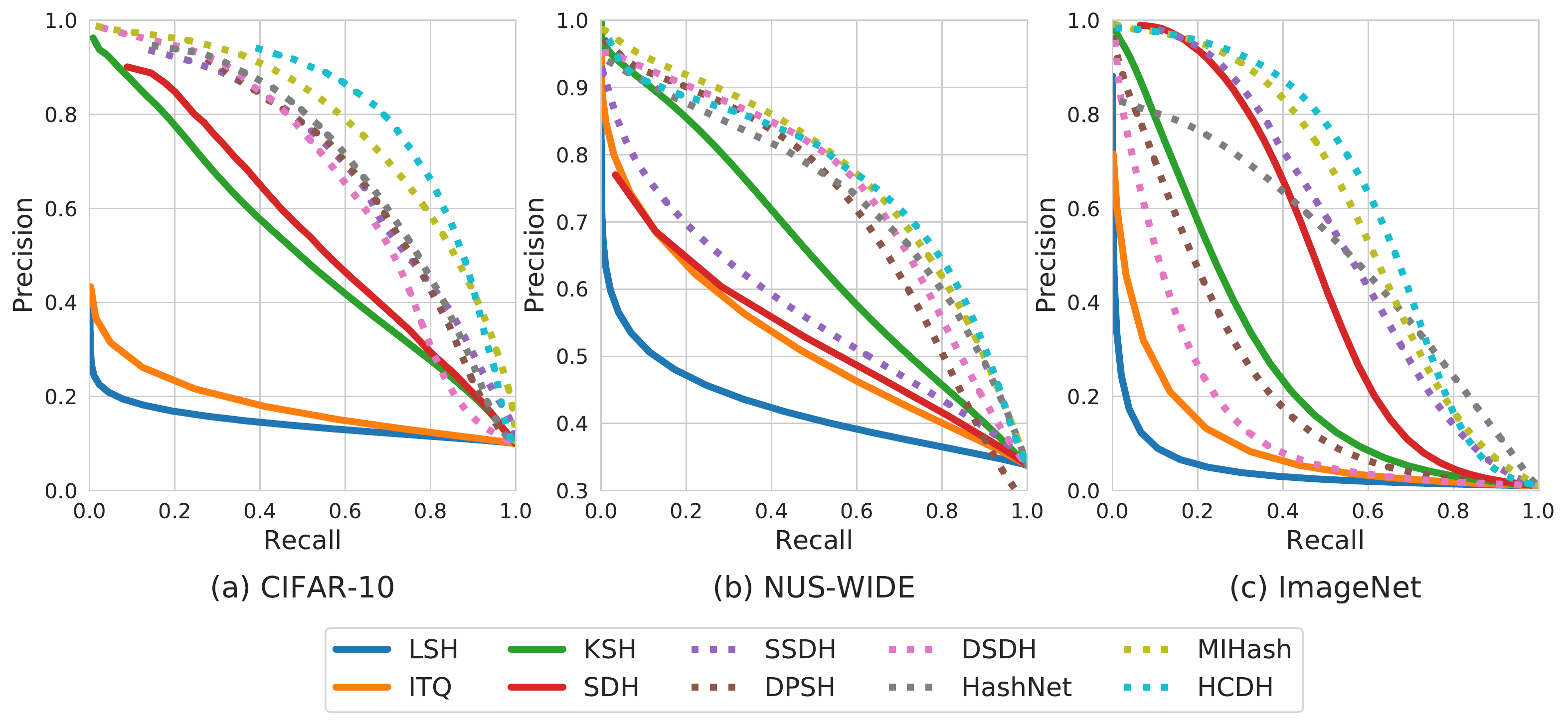}
		\caption{Precision-Recall curves on three datasets. The code length is 64.}
		\label{precision-recall}
	\end{figure*}
	%%%%%%%%%%%%%%%%%%%%%%%%%%%%%%%%%%%
	
	\section{Experiments}
	\subsection{Datasets and Evaluation setup}
	We conduct extensive evaluations on three widely-used benchmark datasets, \emph{i.e.}, CIFAR-10~\cite{krizhevsky2009learning}, NUS-WIDE~\cite{Chua2009NUSWIDEAR}, and ImageNet~\cite{Deng2009ImageNetAL}. 
	
	\begin{itemize}
		\item \textbf{CIFAR-10} is a dataset containing 60,000 images evenly divided into 10 categories. Following the protocol in~\cite{Lai2015SimultaneousFL}, we randomly select 100 images per class as the query set, 500 images per class as the training set, and the rest are used to form the database.
		\item \textbf{NUS-WIDE} is a dataset which contains 269,648 images in 81 ground truth categories. Following the protocol in~\cite{Lai2015SimultaneousFL}, we consider a subset of 195,834 images associated with the 21 most frequent concepts, and randomly sample 100 images per class to form the query set, 500 images per class to form the training set. The remaining forms the database.
		\item \textbf{ImageNet} is a dataset containing over 1.2M images in the training set and 50K images in the validation set, where each image is single-labeled by one of the 1,000 categories. Following the protocol in~\cite{cao2017hashnet}, we randomly select 100 categories, use all the images of these categories in the training set to form the database, and use all the images in the validation set to form the query set. We also randomly select 100 images per category from the database to form the training set.
	\end{itemize}
	
	We evaluate the retrieval results based on two widely-adopted metrics: mean Average Precision ($m$AP) and Precision-Recall curves (PR curves). Following the protocol in~\cite{cao2017hashnet}, we adopt $m$AP@5000 and $m$AP@1000 for NUS-WIDE and ImageNet datasets due to their large scale, respectively.

	We compare the retrieval performance of our method with several classic non-deep hashing methods such as LSH~\cite{Gionis1999SimilaritySI}, ITQ~\cite{Gong2013IterativeQA}, KSH~\cite{Liu2012SupervisedHW}, SDH~\cite{Shen2015SupervisedDH}, and the state-of-the-art deep hashing methods including SSDH~\cite{Yang2015SupervisedLO}, DPSH~\cite{Li2015FeatureLB}, DSDH~\cite{Li2017DeepSD}, HashNet~\cite{cao2017hashnet} and MIHash~\cite{cakir2019hashing}.
	For deep hashing methods, we directly use the raw image pixels as the inputs and adopt the AlexNet~\cite{Krizhevsky2012ImageNetCW} as the backbone network. For non-deep hashing methods, we use the deep features extracted from the AlexNet pre-trained on ImageNet as inputs.
	To guarantee a fair comparison, the results of baselines are executed using the implementations kindly provided by their authors. 
	
	We implement the HCDH on open-source PyTorch~\cite{Paszke2017AutomaticDI}. The parameters of network are initialized by the pre-trained AlexNet on ImageNet. In the training phase, we employ stochastic gradient descent (SGD) with 0.9 momentum and 0.0005 weight decay, and set the min-batch size to 128. The learning rate is set to an initial value of $10^{-4}$, with 50\% decrease every 50 epochs. The weight parameter $\lambda$ for three datasets are empirically set to 1, 0.1 and 0.01, respectively.

	\subsection{Results and Discussions}
	Tab.\,\ref{mAP} shows the $m$AP comparisons on CIFAR-10, NUS-WIDE and ImageNet with respect to different bits number. We observe that: 
	(1) HCDH substantially outperforms all comparison methods. For example, compared with the state-of-the-art, MIHash, HCDH improves the average $m$AP on CIFAR-10 and ImageNet by 3.5\% and 4.2\%, respectively. Similar results can be observed on the multi-label dataset, NUS-WIDE. In addition, HCDH performs very well even in low bits (\emph{e.g}, 16 bits), while other methods show a significant decrease. To explain, HCDH adopts Hadamard codebook to guide the learning of hash codes. Since the balance and independence are guaranteed by Hadamard codebook, the information gain of each bit is maximized, which ensures our excellent performance in low bits. In comparison, the affinity matching approaches (\emph{e.g.}, MIHash, HashNet) require a longer length of bits to achieve similar results. 
	(2) Compared with SSDH that includes the balance and independence in the loss design, HCDH has a considerable improvement on all three datasets, which indicates that the numerical optimization is hard to find a global minimum. Instead, it introduces the guidance from Hadamard codebook to make a big difference. 
	(3) The pairwise-based DPSH and DSDH show poor performance on ImageNet. This is due to the data imbalance problem between similar and dissimilar pairs~\cite{cao2017hashnet,cao2018deep}. Differently, HCDH is trained in a point-wise manner and is not affected by the data imbalance. Hence, our method is more robust and suitable for large-scale datasets. 
	(4) In most cases, the deep hashing methods perform better than the traditional hashing methods, which indicates the effectiveness of incorporating deep neural network in hashing.
	
	Fig.\,\ref{precision-recall} shows the retrieval performance in terms of Precision-Recall curves (PR curves) with respect to different bits number. HCDH delivers higher precision than the state-of-the-arts at the same recall rate on both CIFAR-10 and ImageNet. Competitive results can also be observed on NUS-WIDE. This further demonstrates that HCDH is also favorable for precision-oriented retrieval systems.
	
	\subsection{Visualization of Hash Codes}
\begin{figure}[t]
		\centering
		\subfigure[HCDH]{\includegraphics[width=0.23\textwidth]{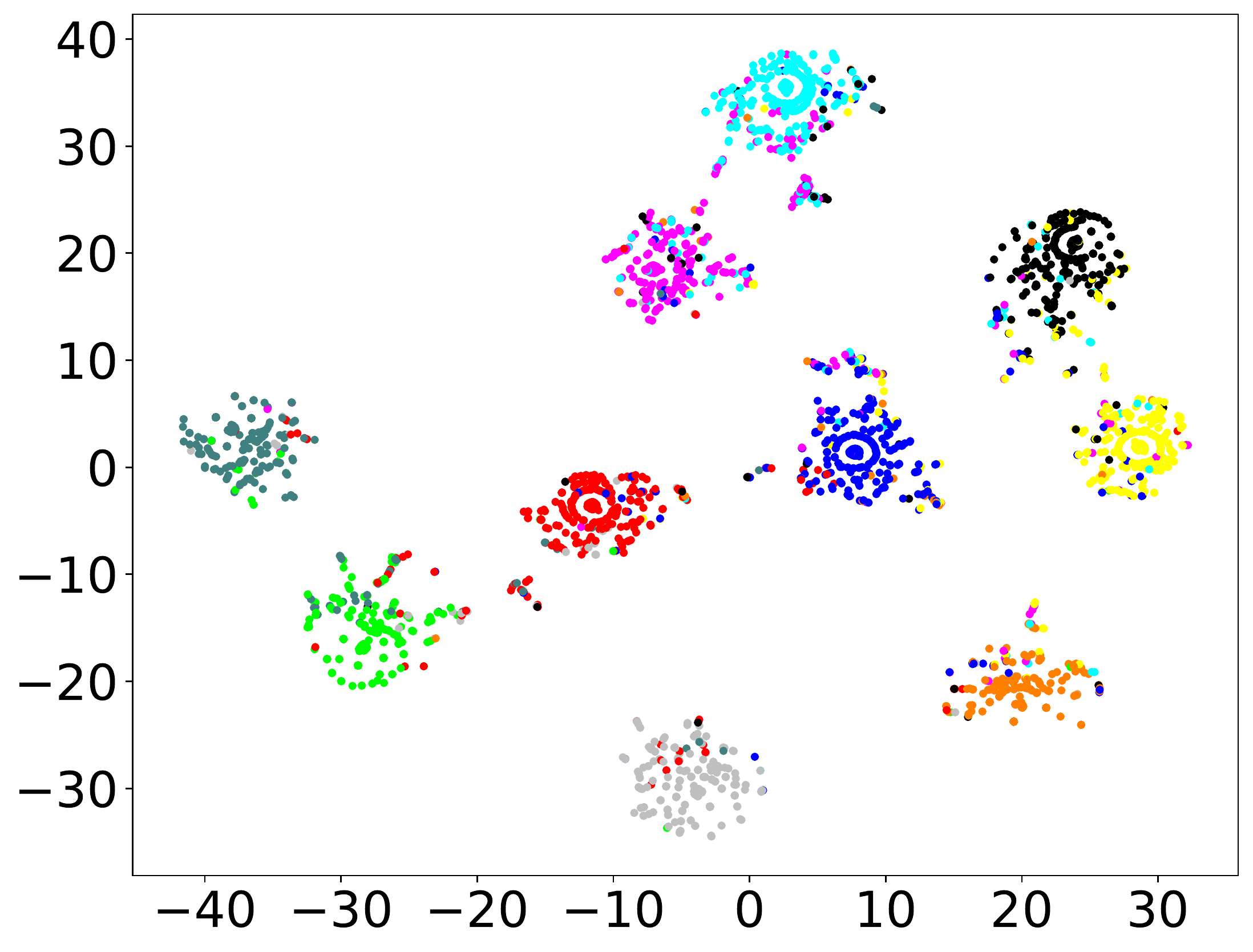}}
		\subfigure[MIHash]{\includegraphics[width=0.23\textwidth]{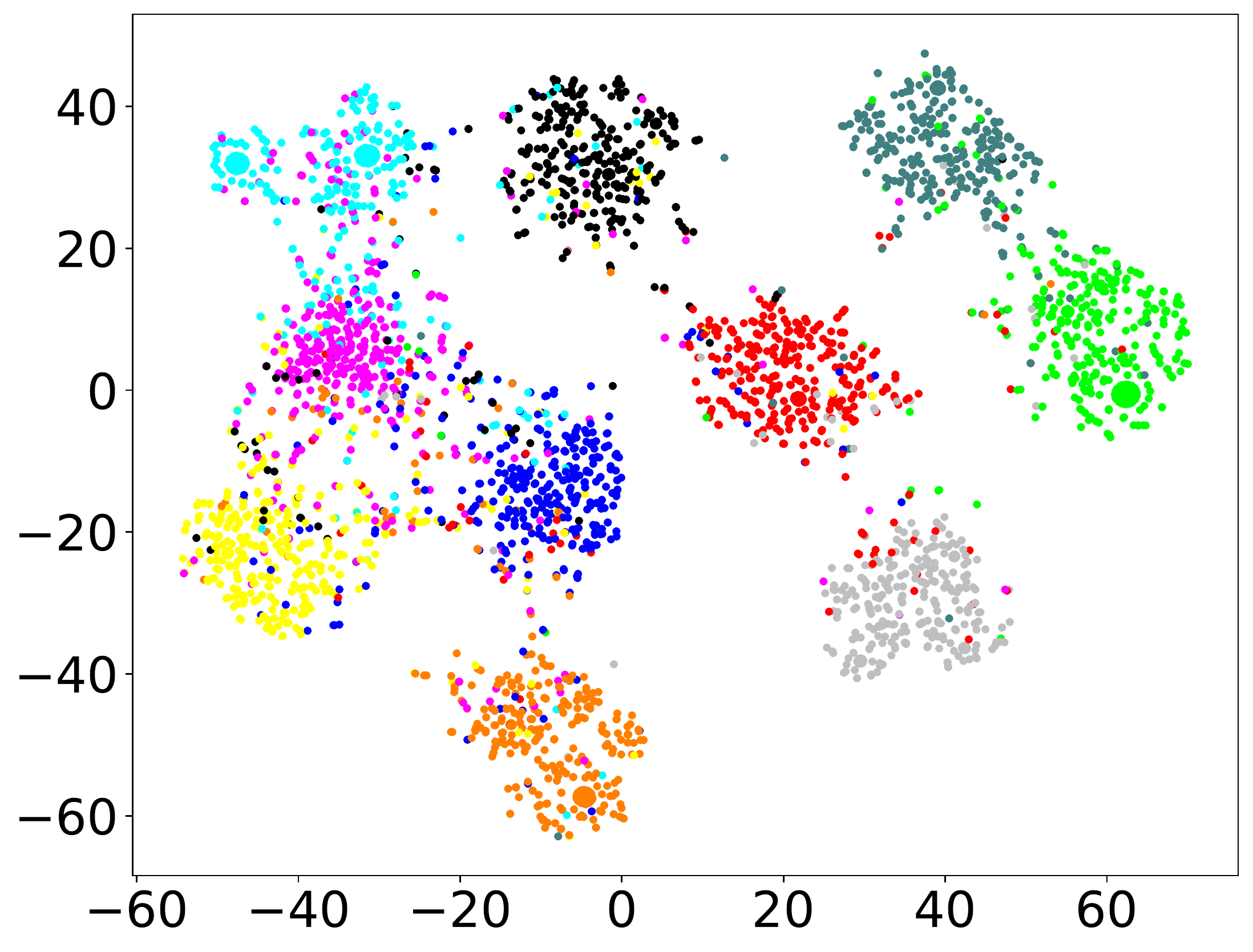}}
		\caption{The t-SNE of hash codes learned by HCDH and MIHash on CIFAR-10.}
		\label{t-SNE}
	\end{figure}

	We visualize the learned embeddings using t-SNE~\cite{maaten2008visualizing}. As illustrate in Fig.\,\ref{t-SNE}, we plot the visualization for 64-bits hash codes produced by HCDH and the top competing method, \emph{i.e.}, MIHash, on CIFAR-10. On one hand, the hash codes generated by HCDH show discriminative structures among different classes. This is indeed predictable from the Hadamard codebook of HCDH, in which the codeword of each class is orthogonal to each other. On the other hand, hash codes generated by MIHash have higher overlap between classes. This is also consistent with the fact that MIHash does not specifically optimize for a criterion related to class overlap, which belongs to the simpler affinity matching approaches.
	
	%%%%%%%%%%%%%%%%%%%%%%%%%%%%%%%%%%%w
	%%%%%%%%%%%%%%%%%%%%%%%%%%%%%%%%%%%
	
	%%%%%%%%%%%%%%%%%%%%%%%%%%%%%%%%%%%w		
	\begin{figure}[t]
		\centering
		\subfigure[HCDH]{\includegraphics[width=0.23\textwidth]{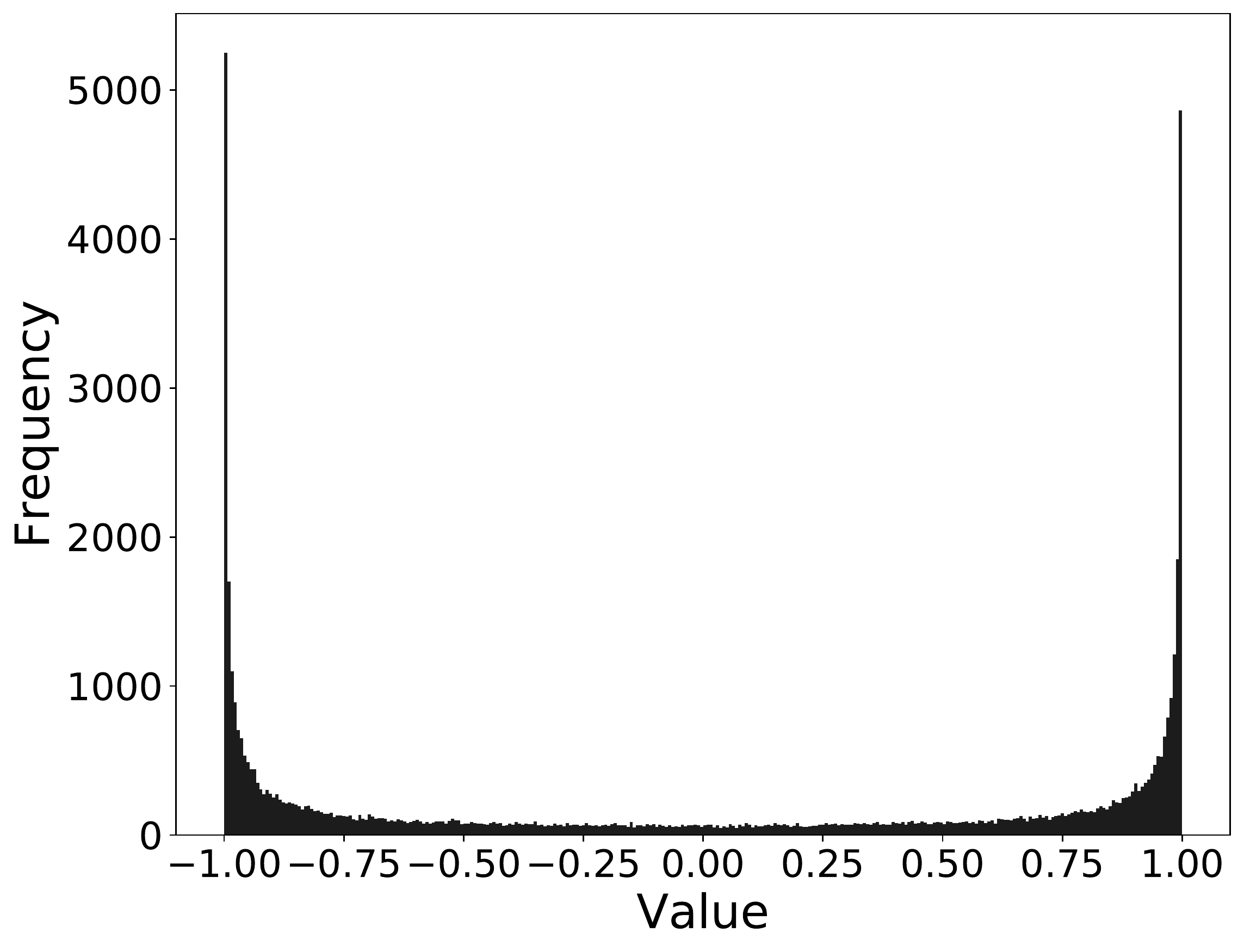}}
		\subfigure[MIHash]{\includegraphics[width=0.23\textwidth]{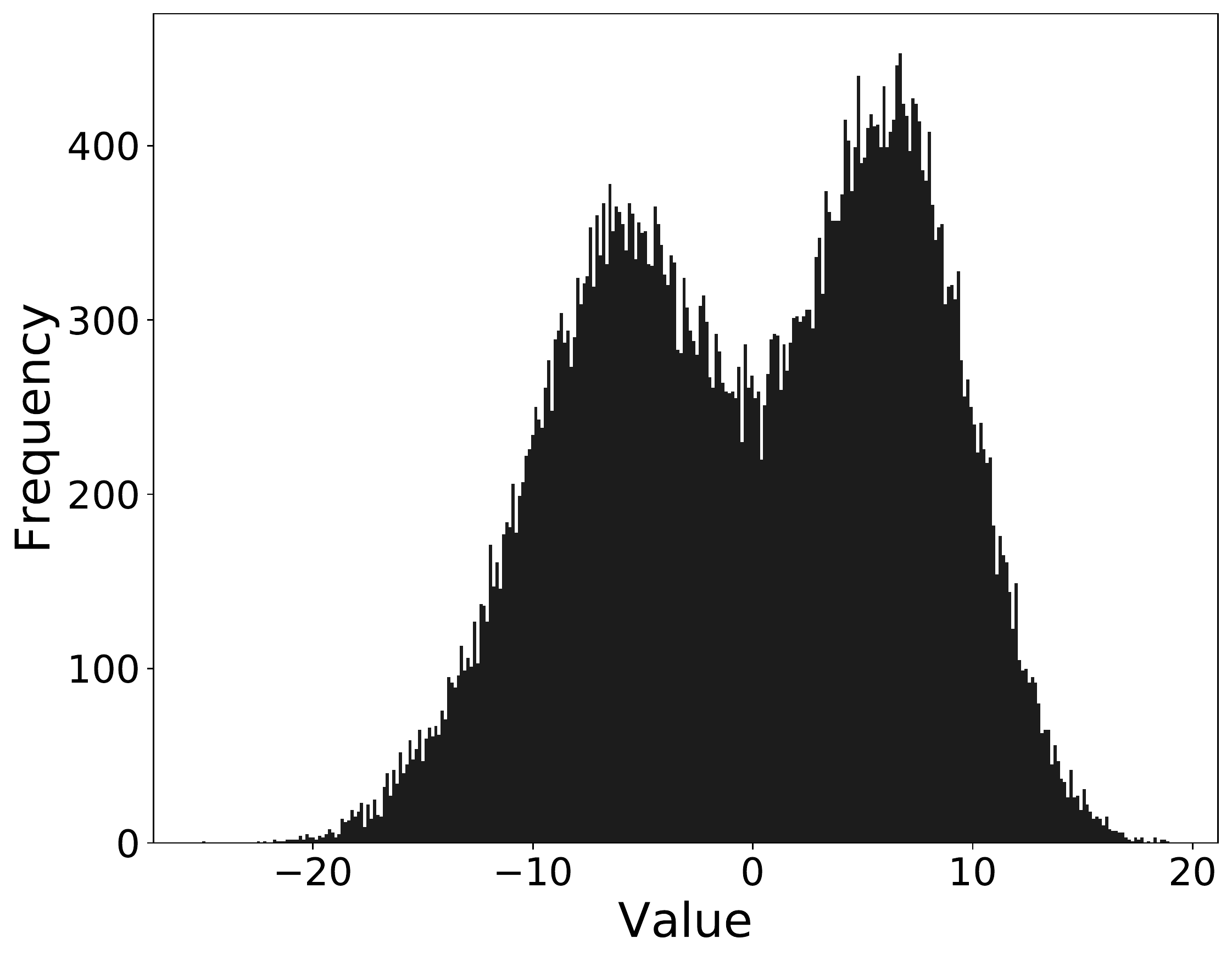}}
		\caption{The distribution of hash features (without $sgn$) learned by HCDH and MIHash on CIFAR-10.}
		\label{distribution}
	\end{figure}
	%%%%%%%%%%%%%%%%%%%%%%%%%%%%%%%%%%%
	
	%%%%%%%%%%%%%%%%%%%%%%%%%%%%%%%%%%%
	\begin{figure}[t]
		\centering
		\includegraphics[width=0.35\textwidth]{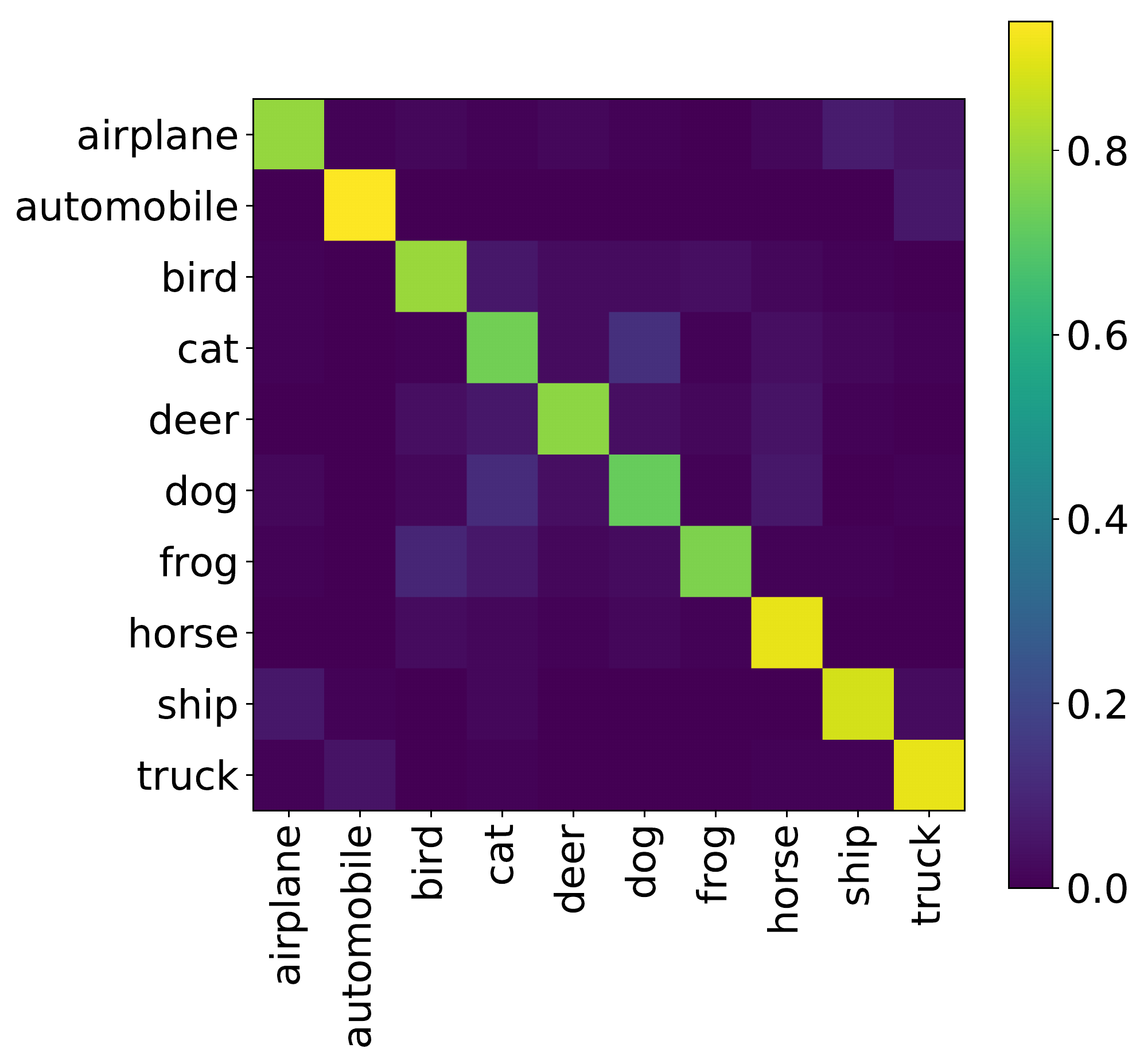}
		\caption{The confusion matrix on CIFAR-10.}
		\label{confusion-matrix}
	\end{figure}
	%%%%%%%%%%%%%%%%%%%%%%%%%%%%%%%%%%%
	
	%%%%%%%%%%%%%%%%%%%%%%%%%%%%%%%%%%%
	\begin{figure}[t]
		\centering
		\includegraphics[width=0.4\textwidth]{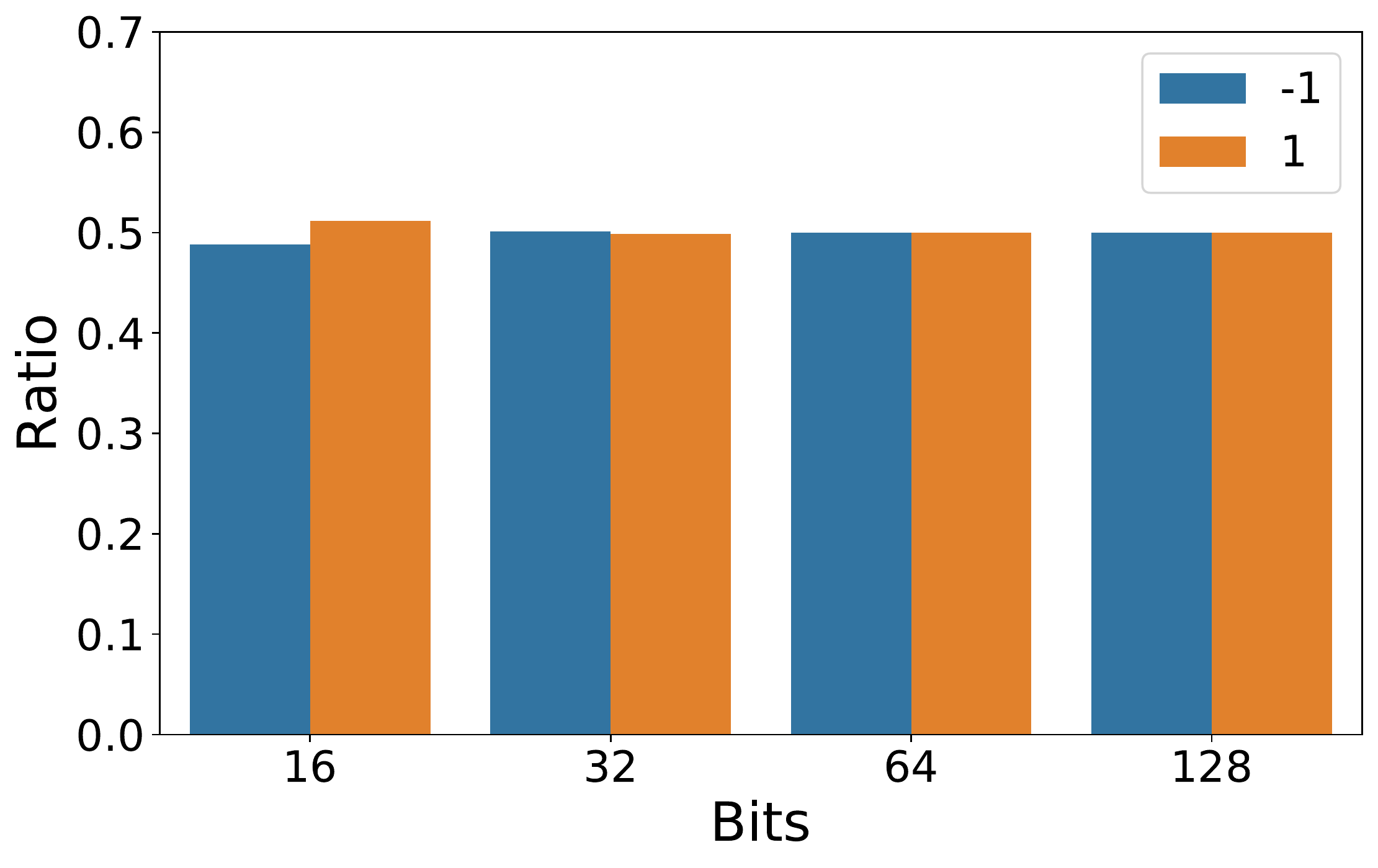}
		\caption{The ratio of $+1s$ and $-1s$ with respect to different bits number in the hash codes learned by HCDH on CIFAR-10.}
		\label{Bihistogram}
	\end{figure}
	%%%%%%%%%%%%%%%%%%%%%%%%%%%%%%%%%%%
	
	%%%%%%%%%%%%%%%%%%%%%%%%%%%%%%%%%%%
	\begin{table*}[t]
		\centering
		\caption{$m$AP results of HCDH and its variants, HCDH-H, HCDH-C, and HCDH-2 on three datasets.}
		\label{$m$AP-variants}
		\setlength{\tabcolsep}{4.9pt}
		\begin{tabular}{c|cccc|cccc|cccc}
			\hline
			\multirow{2}{*}{Method} & \multicolumn{4}{c|}{CIFAR-10} & \multicolumn{4}{c|}{NUS-WIDE} & \multicolumn{4}{c}{ImageNet}\\
			\cline{2-13}
			& 16 bits & 32 bits & 64 bits & 128 bits & 16 bits & 32 bits & 64 bits & 128 bits & 16 bits & 32 bits & 64 bits & 128 bits \\
			\hline
			HCDH 			&\textbf{0.769}   &\textbf{0.774}   &\textbf{0.785}   &\textbf{0.790}   &\textbf{0.779}   &\textbf{0.820}   &\textbf{0.830}   &\textbf{0.832}   &\textbf{0.636}   &\textbf{0.691}   &\textbf{0.719}   &\textbf{0.732}   \\
			HCDH-H			&0.749   &0.759   &0.756   &0.740   &0.772   &0.812   &0.823   &0.827   &0.609   &0.681   &0.708   &0.711   \\
			HCDH-C			&0.740   &0.751   &0.767   &0.768   &0.625   &0.734   &0.784   &0.798   &0.616   &0.678   &0.694   &0.700   \\
			HCDH-2			&0.733   &0.744   &0.755   &0.764   &0.771   &0.813   &0.821   &0.825   &0.600   &0.665   &0.690   &0.704   \\
			\hline
		\end{tabular}
	\end{table*}
	%%%%%%%%%%%%%%%%%%%%%%%%%%%%%%%%%%%
	
	%%%%%%%%%%%%%%%%%%%%%%%%%%%%%%%%%%%
	\begin{figure}[t]
		\centering
		\includegraphics[width=0.45\textwidth]{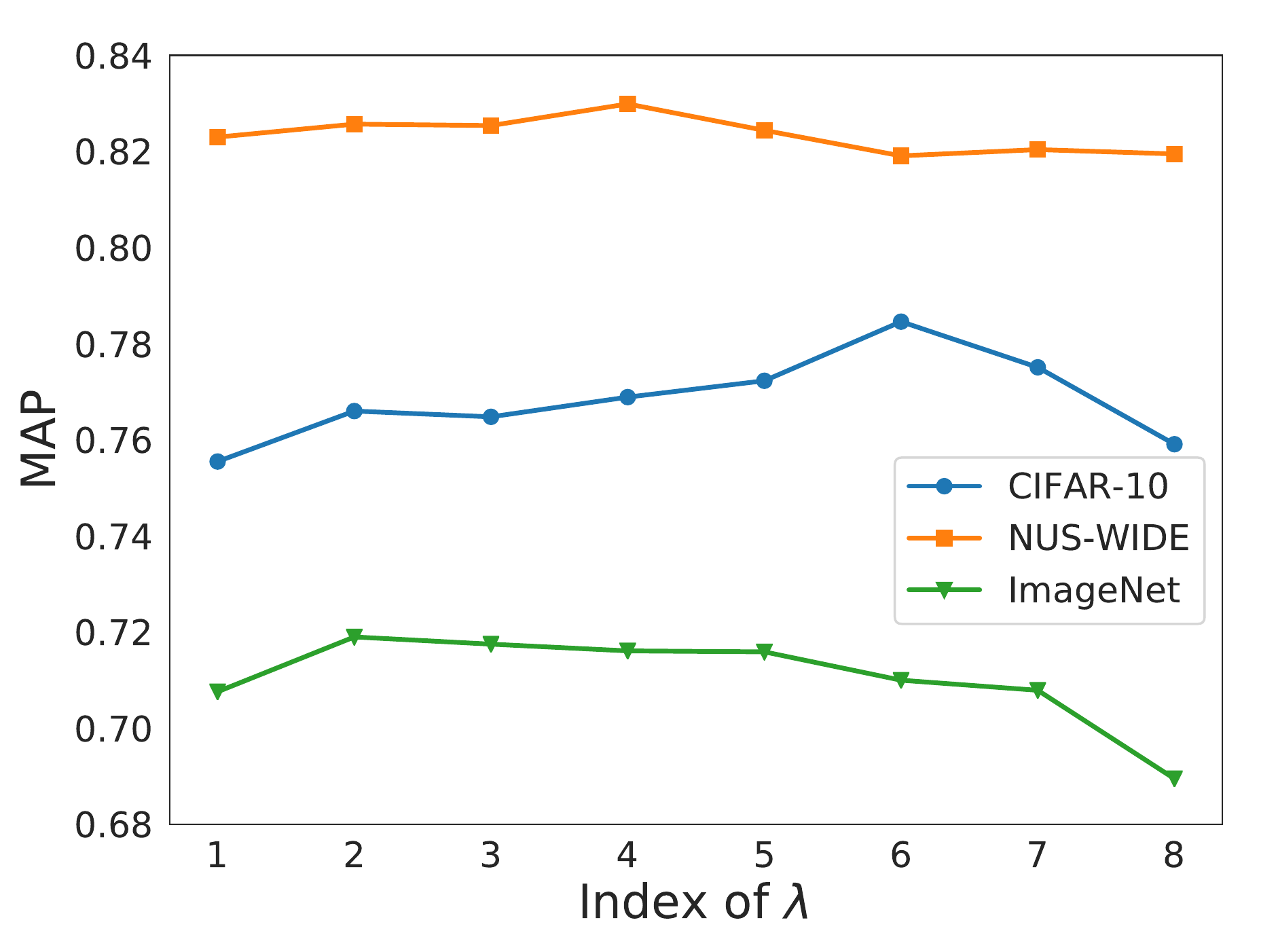}
		\caption{$m$AP with respect to different $\lambda$ on three datasets. The value $\lambda$ is selected from [0, 0.01, 0.05, 0.1, 0.5, 1, 5, 10] and the code length is 64.}
		\label{params-sensitivity}
	\end{figure}
	%%%%%%%%%%%%%%%%%%%%%%%%%%%%%%%%%%%
	
	\subsection{Analysis of Hash Properties}
	To further demonstrate the effectiveness of Hadamard codebook for hashing, we analyze how HCDH ensures three key properties of hash codes, \emph{i.e.}, binarization, independence and balance.
	
	\subsubsection{Binarization.}
	To illustrate the binarization of Hadamard codebook, Fig.\,\ref{distribution} presents the distribution of hash features (without $sgn$) produced by HCDH and MIHash on CIFAR-10. The code length is set to 64. Clearly, we can see that the distribution of HCDH is almost concentrated around 1 and -1, while MIHash shows a bimodal distribution and concentrates around 0.
	This is due to the fact that HCDH adopts the binary Hadamard codebook as guidance information, which directly pushes hash codes toward 1 and -1 in the training stage.
	In contrast, MIHash can be regarded as a sort of embedding learning, which ignores the essential properties of hashing, \emph{e.g.}, binarization, and inevitably leads to \textit{quantization error}~\cite{Gong2013IterativeQA}.
	
	\subsubsection{Independence.}
	To illustrate the independence of Hadamard codebook, Fig.\,\ref{confusion-matrix} presents the confusion matrix produced by HCDH on CIFAR10. Specifically, for each query point, we calculate the frequency of different classes according to the retrieval results and give the tops larger weights to obtain the desired confusion matrix. An entry with higher brightness indicates that the corresponding class is retrieved more correctly, and vice versa. It is obvious that the diagonal entries of the confusion matrix are the brightest, while the rest are mostly close to 0. This is mainly due to the orthogonality of Hadamard codebook, which ensures the distinction between classes. Besides, the entry values between similar classes are slightly higher, \emph{e.g.}, cat and dog, which indicates that the semantic information between classes is also well preserved.
	
	\subsubsection{Balance.}
	To illustrate the balance of Hadamard codebook, we calculate the ratio of $+1s$ and $-1s$ with respect to different bits number in the hash codes generated by HCDH on CIFAR-10. The results are shown in Fig.\,\ref{Bihistogram}. It's very clear that the number of $+1s$ and $-1s$ in the hash codes is basically the same across all the bits. It validates that HCDH can learn balanced hash codes, which maximizes the information gain in each bit.
	
	\subsection{Parameter Sensitivity}
	
	The $m$AP results of HCDH with respect to different values of the hyper-parameter $\lambda$ on three datasets are shown in Fig.\,\ref{params-sensitivity}. We tune the value of $\lambda$ in the range of [0, 0.01, 0.05, 0.1, 0.5, 1, 5, 10], and set the code length to 64. By imposing a large $\lambda$, \emph{e.g.}, close to 10, the HCDH gradually degenerates into a simple classification model. Due to the lack of guidance from Hadamard codebook, the $m$AP results on three datasets have a significant decrease. By imposing a small $\lambda$, \emph{e.g.}, close to 0, HCDH merely utilizes the Hadamard codebook to learn hash codes. As can be seen from the experimental results, the retrieval performance first ascends and then decreases. The best performances for CIFAR-10, NUS-WIDE and ImageNet are obtained when the values of $\lambda$ are set to 1, 0.1, and 0.01, respectively. 
	
	\subsection{Abalation Study}
	To evaluate the contributions of Hadamard codebook and the co-trained deep classifier on the final performance, we investigate three variants of HCDH: (1) \textbf{HCDH-H}, a variant only using Hadamard codebook for training; (2) \textbf{HCDH-C}, a variant only using deep classifier for training; (3) \textbf{HCDH-2}, the variant adopting the two-stream architecture~\cite{Yang2015SupervisedLO} instead of our one stream architecture. The $m$AP results with respect to different bits number on three benchmarks are reported in Tab.\,\ref{$m$AP-variants}.
	
	By exploiting semantic information via deep classifier, HCDH outperforms HCDH-H by 2.9\%, 0.7\% and 1.7\% respectively in average $m$AP. We attribute this to the random selection of Hadamard codebook from Hadamard matrix, which cannot guarantee the semantic similarity between classes. Similarly, HCDH-C suffers from an average $m$AP decreases of 2.3\%, 8.0\% and 2.3\%, especially on NUS-WIDE, which substantially underperforms HCDH. These results show that using only the classification model can not ensure the discriminability of hash codes, and is not suitable for multi-label datasets in practice. It is worth noting that, in most cases, HCDH-H outperforms HCDH-C, which demonstrates the superiority of the Hadamard codebook in hash learning.
	
	Another key observation is, by using two-stream architecture, HCDH-2 incurs large average $m$AP decreases of 3.1\%, 0.8\% and 3.0\% compared with HCDH. In the two-stream framework, the classification stream is only employed to learn image representation, which does not contribute directly to the learning of hash functions. In contrast, HCDH uses CNN to learn the image representation and hash functions simultaneously. The hash codes are directly guided by the Hadamard codebook and classification information.
	
	\section{Conclusion}
	In this paper, we propose a novel deep supervised hashing method, called HCDH, for large-scale image retrieval. With the power of Hadamrd codebook, the issues of bit independence and bit balance in the existing deep hashing methods can be effectively solved. We also introduce a deep classifier to further exploit the supervised labels. Comprehensive experiments justify that HCDH generates balanced and discriminative binary codes that yield state-of-the-art performance on three standard benchmarks, \emph{i.e.}, CIFAR-10, NUS-WIDE, and ImageNet.
	
	\bibliography{mybibliography}
	\bibliographystyle{aaai}
\end{document}